\documentclass[lettersize,journal]{IEEEtran}
\usepackage{amsmath,amsfonts}
\usepackage{array}
\usepackage[caption=false,font=normalsize,labelfont=sf,textfont=sf]{subfig}
\usepackage{textcomp}
\usepackage{stfloats}
\usepackage{url}
\usepackage{verbatim}
\usepackage[noend]{algpseudocode}  
\usepackage[]{algorithm2e}
 \usepackage[switch]{lineno}
\usepackage{graphicx}
\usepackage{multirow}
\usepackage[table,xcdraw]{xcolor}
\usepackage[graphicx]{realboxes}
\usepackage{color}
\usepackage{cite}
\usepackage{xcolor}
\usepackage[colorlinks,
            linkcolor=blue,       
            anchorcolor=blue,  
            citecolor=blue,        
            ]{hyperref}
\usepackage{utfsym}
\usepackage[normalem]{ulem}
\useunder{\uline}{\ul}{}

\begin{document}

\title{Optimizing 4D Lookup Table for Low-light Video Enhancement via Wavelet Priori}




%
\author{Jinhong He, Minglong Xue*, Wenhai Wang and Mingliang Zhou
\thanks{*Corresponding author: Minglong Xue.
Jinhong He and Minglong Xue are with the College of Computer Science and Engineering, Chongqing University of Technology, Chongqing, 400054, China (e-mail: hejh@stu.cqut.edu.cn, xueml@cqut.edu.cn );
Wenhai Wang is with the Chinese University of Hong Kong, Hong Kong, China (e-mail: whwang@ie.cuhk.edu.hk);
Mingliang Zhou is with the College of Computer Science, Chongqing
University, Chongqing 400044, China (e-mail:mingliangzhou@cqu.edu.cn).



}
}


\maketitle

\begin{abstract}
Low-light video enhancement is highly demanding in maintaining spatiotemporal color consistency. Therefore, improving the accuracy of color mapping and keeping the latency low is challenging. Based on this, we propose incorporating Wavelet-priori for 4D Lookup Table (WaveLUT), which effectively enhances the color coherence between video frames and the accuracy of color mapping while maintaining low latency. Specifically, we use the wavelet low-frequency domain to construct an optimized lookup prior and achieve an adaptive enhancement effect through a designed Wavelet-prior 4D lookup table. To effectively compensate the a priori loss in the low light region, we further explore a dynamic fusion strategy that adaptively determines the spatial weights based on the correlation between the wavelet lighting prior and the target intensity structure. In addition, during the training phase, we devise a text-driven appearance reconstruction method that dynamically balances brightness and content through multimodal semantics-driven Fourier spectra. Extensive experiments on a wide range of benchmark datasets show that this method effectively enhances the previous method's ability to perceive the color space and achieves metric-favorable and perceptually oriented real-time enhancement while maintaining high efficiency. 
\end{abstract}


\section{Introduction}
\IEEEPARstart{R}{eal}-world low-light environmental conditions often result in severe degradation of the quality of the captured video. Low-light video enhancement (LLVE) aims to reverse the degraded domain and improve the visibility and visual quality of videos captured in low-light conditions, which is important for a variety of downstream tasks, such as automatic driving \cite{li2024light}, unmanned aerial vehicle navigation \cite{ye2021darklighter}, text detection \cite{xue2020arbitrarily} and Photography on mobile devices \cite{ignatov2017dslr}. In hardware-based approaches, researchers usually use high ISO, long exposure time, and large aperture for video enhancement. However, these methods have limitations\cite{chen2018learning,cheng2016learning}. For example, high ISO is limited because it can amplify noise, and long exposure time can lead to motion blur. On the other hand, along with the rise of deep learning, numerous researchers have utilized raw videos to provide degraded a priori knowledge for low-light enhancement through a data-driven approach \cite{jiang2019learning,chen2019seeing,wang2021seeing,fu2023dancing}.

In recent years, depth models have gradually dominated video processing tasks. In addition, video tasks are different from image tasks in that they are strongly correlated in time and space and require effective stability to ensure consistent color brightness in time. Therefore, directly applying image-based methods \cite{xu2022snr,liang2023iterative} in video tasks is not reliable. This will lead to the emergence of the video flicker problem \cite{chen2019seeing,li2023fastllve} and increase the consumption of computational resources to some extent. Therefore, researchers have mitigated the inter-frame brightness variations by exploring the spatio-temporal information of the video. For example, SMOID \cite{jiang2019learning} utilizes 3D convolution to aggregate temporal features. StableLLVE \cite{zhang2021learning}chose to use optical flow to align adjacent frame features. However, these approaches do not strike a balance between performance and efficiency, increasing the challenge of real-time applications in the real world.

\begin{figure}[t]
        \centering
        \includegraphics[height=0.2\textwidth]{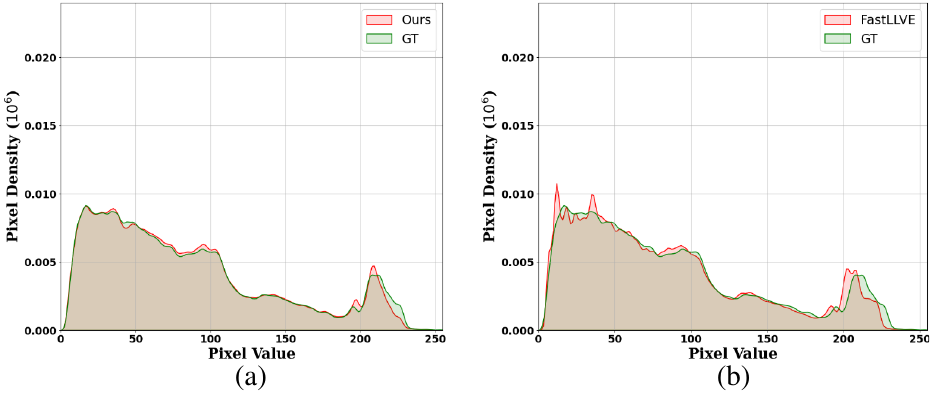}
        \caption{Comparison with FASTLLVE on the pixel distribution of the enhancement results. It can be seen that we mapped pixel values closer to the reference video.}
        \label{fig:1}
    \end{figure}
    
\begin{figure}[t]
        \centering
        \includegraphics[height=0.24\textwidth,width=0.48\textwidth]{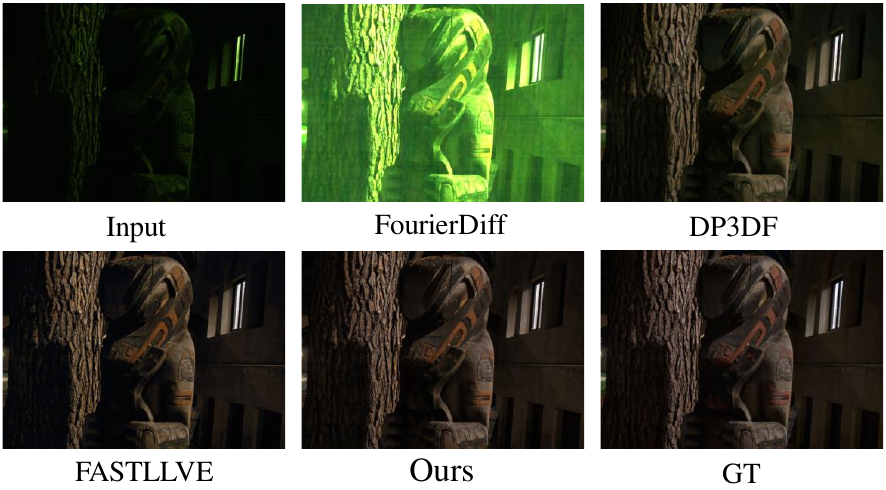}
        \caption{Our method is compared with SOTA method DP3DF \cite{xu2023deep}, FASTLLVE\cite{li2023fastllve}. We can achieve more accurate color mapping, resulting in a friendlier visual effect.}
        \label{fig:lv}
    \vspace{-10pt}
    \end{figure}

Due to the efficient modeling of Lookup Table (LUT) and more subtle color mapping, researchers have introduced it into video tasks to alleviate the above problems. For example, \cite{wang2021real} constructed a spatial-aware LUT by combining global scene and local spatial information. FASTLLVE \cite{li2023fastllve} combines intensity maps to construct an intensity-aware LUT for subtle color transformations. However, despite the efforts of previous methods in constructing lookup priors to mimic the underlying optimal color transformations, these lookup priors are still corrupted to a certain extent due to the complexity and variability of low-light environments, as shown in Fig. \ref{fig:1}, resulting in more deviations of pixel values in the color mapping space. And Fig. \ref{fig:lv} also visually compares the enhanced colors.

\begin{figure}[t]
        \centering
        \includegraphics[height=0.35\textwidth,width=0.48\textwidth]{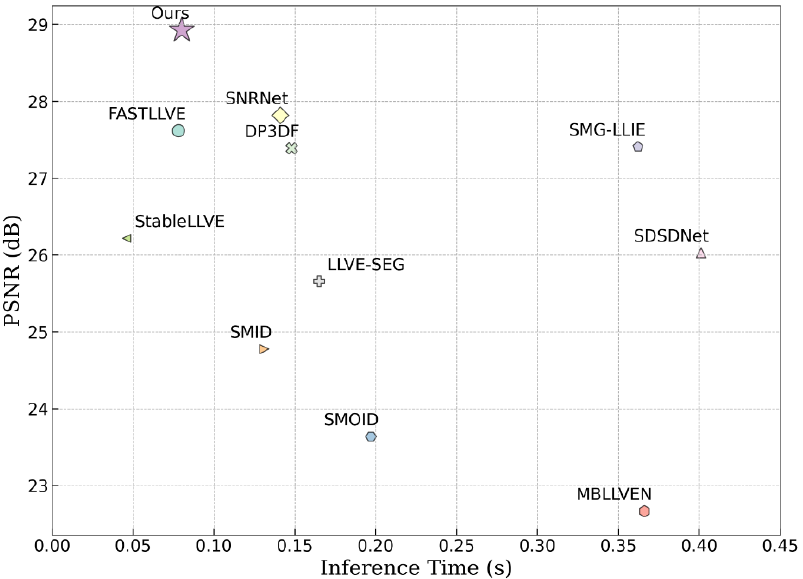}
        \caption{Based on the SMID dataset, we use the Nvidia RTX 3090 GPU to target $1920\times1080$ (1080p) video comprehensively comparing PSNR and efficiency. Our approach has a better balance in the evaluation of performance and efficiency.}
        \label{fig:2}
    \vspace{-1em}
    \end{figure}

To further address the above issues, this paper proposes incorporating wavelet prior to the 4D Lookup Table (WaveLUT), enhancing the accuracy of color space mapping while maintaining efficiency. Specifically, as revealed in \cite{cai2023retinexformer}, a rough low-light prior facilitates image recovery and further improves performance. Therefore, we extract the low-frequency domain of coherent low-light video data via wavelet transform to generate lighting prior for fusion to compensate for the damaged lookup table prior in low-light environments. Subsequently, to generate the optimal colors at the bottom of the lookup table and to match the constructed lookup prior, we combine the wavelet low-frequency domain and follow the parameterization approach \cite{zeng2020learning} to construct the Wavelet-prior 4D LUT via a set of basis 4DLUTs.  Moreover, to construct the lookup prior efficiently, we further explore a dynamic fusion strategy to enhance the lookup prior by calculating the spatiotemporal correlation between the lighting and intensity prior to dynamically fusing different content structures. During the training phase, we also devise a text-driven appearance reconstruction method that dynamically balances brightness and content by combining multimodal semantics and Fourier spectra to bring the enhancement results closer to the reference video. Extensive experiments on benchmark datasets validate the effectiveness of our method. As shown in Fig. \ref{fig:2}, our method maintains efficient processing efficiency and is more competitive in peak signal-to-noise ratio (PSNR).

In summary, the contribution of this paper can be summarised as follows:
\begin{itemize}
\item We propose incorporating wavelet priori for the 4D Lookup Table (WaveLUT) method. It effectively enhances the mapping accuracy of the lookup table while maintaining efficiency, which leads to a more friendly visual effect.
\item To further exploit the lookup prior, we explore a dynamic fusion strategy that adaptively determines spatial weights to fuse different prior knowledge.
\item We devise a text-driven appearance reconstruction method that combines multimodal semantics and Fourier spectra to dynamically balance brightness and content to promote enhancement results closer to the reference video.
\item Extensive experiments on benchmark datasets validate the effectiveness of our method. Compared to the baseline methods, we can achieve both metric favorable and perceptually oriented enhancements.
\end{itemize}

The remainder of this paper is structured as follows. In Section \ref{Related Works}, the related works are discussed. In Section \ref{Method}, the proposed novel model method is described in detail. The relevant experimental setup and results are shown in Section \ref{EXPERIMENTS}. Section \ref{Conclusions} is the conclusion.

\section{RELATED WORK} \label{Related Works}
\subsection{Low-light Image Enhancement}
Research in low-light image enhancement is gradually gaining attention under the need to mitigate performance degradation in dimly lit environments for numerous vision applications. The original methods are mainly based on traditional optimization methods such as histogram equalization \cite{xu2013generalized,pisano1998contrast} and Retinex theory \cite{gao2017naturalness,zhou}, which are processed using the information of the image itself to produce manual prior. However, due to numerous unknown degradation factors, the high reliance on the production of manual prior does not accurately achieve image optimization. Recently deep learning-based \cite{pan2021miegan,UHDFourICLR2023,xue2024low,li2024perceptual,xu2022illumination,zhou2023low} data-driven approaches have shown better generalization ability and effectiveness. Such as, Zero-DCE \cite{guo2020zero} constructed a pixel-level curve estimation convolutional neural network to achieve efficient enhancement. SNRNet \cite{xu2022snr} designed signal-to-noise ratio-aware transformers and CNN models with spatially varying operations. NeRco \cite{yang2023implicit} introduced a neural normalization approach to keep the degradation level in balance to reduce enhancement difficulty. However, due to the high standard of video on the brightness and color of adjacent frames, the application of image enhancement algorithms to individual frames often suffers from the flicker problem, which leads to the degradation of the visual effect.

\subsection{Low-light Video Enhancement}
To further address the issues arising from low-light image enhancement methods in video tasks \cite{li2023fastllve,zhang2024binarized}, and taking into account the widely popular data formats for videos, researchers extended low-light video enhancement methods \cite{jiang2019learning,zhang2021learning,liang2023coherent,jiang2023video,ni2023online} and video datasets. For video datasets, some work \cite{wei2020physics,wang2021seeing} used camera noise models or Generative Adversarial Networks (GANs) to generate a variety of different low-light videos. LAN \cite{fu2023dancing} reproduces the same motion twice by rigorously enforcing an electromechanical system to obtain paired low/normal light videos of the motion. With the existing video dataset, MBLLEN \cite{lv2018mbllen} replaced the 2D-Conv layer with the 3D-Conv layer for low-light videos. Liu et al. \cite{liu2023low} used event information to learn the enhancement mapping of videos.  DP3DF \cite{xu2023deep} designed a parametric 3D filter to target video data. To improve the efficiency of the network, SMID \cite{chen2019seeing} dealt with small variations between video frames by training a deep twin network based on self-consistency loss. FastDVD \cite{tassano2020fastdvdnet} used a two-stage cascaded U-net for implicit motion compensation. However, current methods have difficulty in refining neighboring pixels in extreme regions in low-light environments, resulting in inaccurate enhancements and requiring more inference time. Also, they ignore the great potential of multimodal semantics in video tasks.

\begin{figure*}[t]
        \centering
        \includegraphics[height=0.38\textwidth,width=\textwidth]{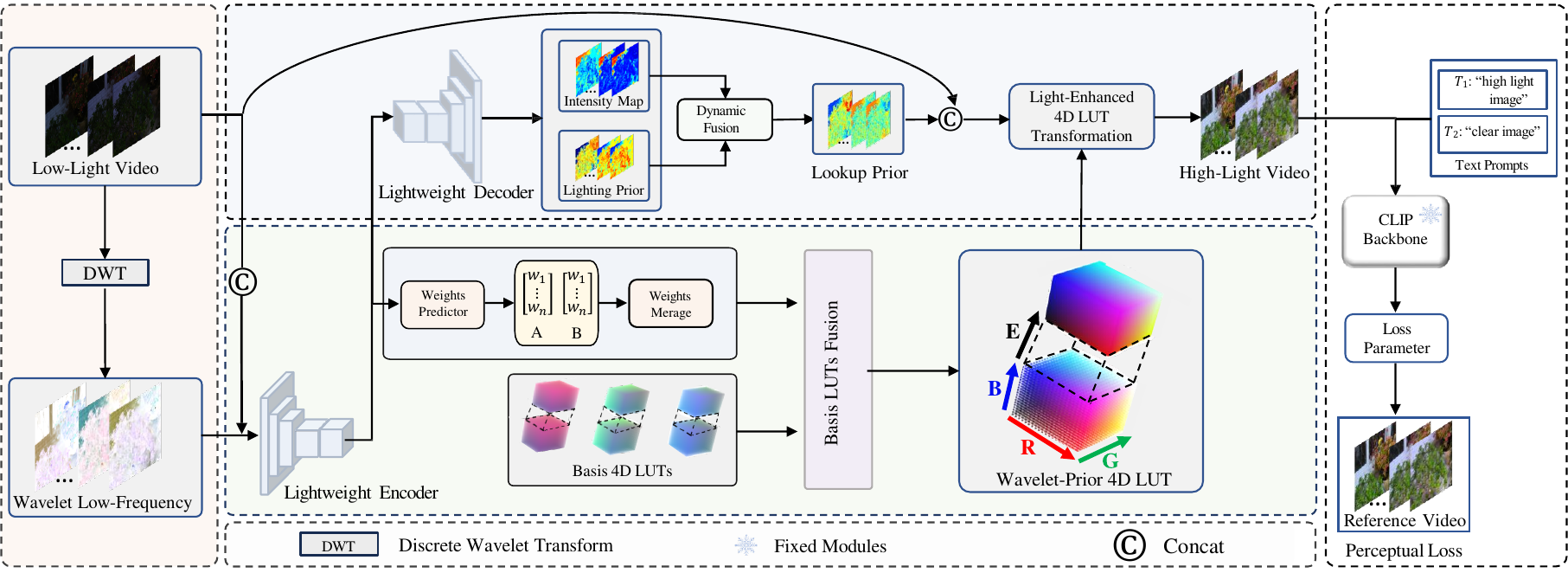}
        \caption{The overall workflow of our proposed WaveLUT. First it extracts the wavelet low-frequency domain of low-light video by discrete wavelet transform $(DWT)$. Subsequently, the wavelet low-frequency domain is used to guide the construction of the Wavelet-prior 4D LUT and optimize the lookup prior in combination with the dynamic fusion strategy. Based on the constructed Wavelet-prior 4D LUT and lookup prior, the input low-light video is converted to enhanced high-light video by light-enhanced 4D LUT transformation. Finally, during the training phase, we further optimize the enhancement results through text-driven appearance reconstruction.}
        \label{fig:7}
    \vspace{-10pt}
    \end{figure*}

\subsection{Lookup Table in Enhancement Tasks}
LUT achieves enhancement by mapping the input color values to the corresponding output color values. LUTs are widely used in image editing software due to their efficient modeling efficiency compared to other methods. In recent years, Zeng et al. \cite{zeng2020learning} first used a lightweight CNN for weight prediction of basis LUT integration and constructed an adaptive LUT for image enhancement. Wang et al. \cite{wang2021real} further learned global scene and local spatial information based on the original foundation. AdaInt \cite{yang2022adaint} designed a learnable non-uniform sampling strategy to alleviate the nonlinear distortion in the transform caused by undersampling of previous methods. Liu et al. \cite{liu20234d} achieved a fine color transformation by extending the 3D LUT to 4D space and adding context-aware images. FASTLLVE \cite{li2023fastllve} then perform real-time video enhancement by constructing an intensity-aware LUT. However, low-light environments cause the constructed lookup prior to being affected to some extent, which makes the mapping more difficult. Therefore, in this paper, we explore using wavelet transform to compensate for the lookup prior and construct 4D LUT for more accurate mapping.


\section{Method} \label{Method}

\subsection{Lookup Tables Preliminaries}
LUT is an efficient color mapping operator. Based on the nature of the three color channels $\{R, G, B\}$ in the image, 3D LUT is usually constructed for color transformation processing in image tasks. 3D LUT is the 3D lattice of values, which maps the input color values to the corresponding output color values and saves them in the 3D lattice $L=\{L_{r,(i,j,k)}, L_{g,(i,j,k)}, L_{b,(i,j,k)}\}_{i,j,k \in S_0^{N-1}}$. $S$ denotes the total set of sampling coordinates for each color channel in RGB, and for each 3D LUT, it defines a total of $N^3$ sampling points to form a complete 3D color transform space. Therefore, each set of input color values mainly relies on the RGB values of the input pixels to obtain the input indexes of $\{ i, j, k\}$ of their color coordinate sets and uses the pixel-to-pixel mapping $\mu(\cdot )$ to find their nearest sample points in the discrete RGB color space. Subsequently, their transformed outputs are computed by trilinear interpolation \cite{liang2021ppr10k}. Applying this transform to an image or video can achieve subtle color gradients to striking color effects:
\begin{equation}
L=\mu \{R_{(i,j,k)} ,G_{(i,j,k)} ,B_{(i,j,k)}\},
\tag{1}
\end{equation}
\begin{small}
\begin{equation}
i=R_{(i,j,k)}\cdot \frac{S}{255},\  j= G_{(i,j,k)}\cdot \frac{S}{255},\  k= B_{(i,j,k)}\cdot \frac{S}{255} .
\tag{2}
\end{equation}
\end{small}

However, constrained by the complexity of low-light environments and the need for spatiotemporal coherence in the video, the one-to-one mapping relation of 3D LUTs does not satisfactorily address pixels with similar colors in dark regions in video enhancement tasks, which makes them suffer from sub-optimal sample point allocation and limited LUT capability. Therefore, inspired by \cite{li2023fastllve}, as shown in Fig. \ref{fig:7},  we implement the one-to-many mapping relation by constructing a 4D LUT based on the wavelet prior. Furthermore, we alleviate the color incoherence difficulties of previous methods and effectively enhance the accuracy of color space mapping. Also, we introduce multimodal semantics to supervise the reconstruction of low-light videos to keep the brightness consistent between video frames.

    
\subsection{4D LUT Construction and Transformation}
\subsubsection{Wavelet-prior 4D LUT} In this paper, we perform a discrete wavelet transform $(DWT)$ on each set of low-light video $V_L\in \mathbb{R}^{N\times H\times W \times C}$ to obtain their wavelet low-frequency domains $V_{WL}=DWT(V_L)$. The $V_{WL} \in \mathbb{R}^{N\times\frac{H}{2}\times\frac{W}{2}\times C}$ contain the illumination information and the main content structure of the low-light videos and also have a rough lighting prior \cite{jiang2023low}. Furthermore, to construct an adaptive Wavelet-prior 4D LUT, we first perform preliminary learning by multiple learnable basis 4D LUTs, and construct video structure-related weights for basis 4D LUT fusion using low-light video and wavelet low-frequency, respectively, by merging the two sets of weights to obtain more video relevance. Specifically, we rely on a lightweight encoder to encode the input $V_L$ and $V_{WL}$ to obtain a rough understanding of the input video and a rough lighting prior. Subsequently, we rely on the resolution of the input video to adjust the two encoding results into feature vectors $Y \in \mathbb{R}^{16\times 64}$ and $Z$ $\in \mathbb{R}^{16\times 64}$, which are used to guide the parameter construction of the video correlation 4D LUT, formulated as:
\begin{equation}
Y,Z=f(\Phi_{E}(V_L),\Phi_{E}(V_{WL})),
\tag{3}
\end{equation}
where $f(\cdot)$ is a function that adjusts the input video to a compact vector, and $\Phi_{E}$ denotes the encoder. Subsequently, we construct a weight predictor through a fully connected layer to map dynamic video-related weights for feature vectors $ Y$ and $Z$, respectively, and based on this, we merge the two weights to enhance the preservation of video information and to construct better underlying colors for the LUT. Finally, we use the merged correlation weights for the fusion of all base 4D LUTs and map them to all elements of the Wavelet-prior 4D LUT through another fully connected layer. The overall process can be described as follows:
\begin{equation}
m_0\begin{Bmatrix}
\Phi_{E}(V_L)\to Y \in \to A \in \mathbb{R}^3  \\
\Phi_{E}(V_{WL})\to Z\in \to B \in \mathbb{R}^3
\end{Bmatrix} \overset{merage}{\rightarrow} \hat{\mathbb{R}}^3  \overset{m_1}{\rightarrow}\mathbb{R}^{N^4\times 3}, 
\tag{4}
\end{equation}
where $m_0$ denotes the mapping from the feature vector to the video correlation weights. $\hat{\mathbb{R}}^3$ denotes the parameters used for basis 4D LUTs fusion after weight merging of the video correlation weights $A$ and $B$. $m_1$ denotes the mapping of the parameters to the Wavelet-prior 4D LUT. $N^4 \times 3$ denotes the total number of elements of the generated Wavelet-prior 4D LUT, where $N$ denotes the number of sampling points on each dimension, and $3$ represents the stored values of the three RGB color mappings, respectively. In addition, since the parameters of the fully connected layer can be updated during training, the underlying 4D LUTs that serve as the parameters can be learned during training. During inference, the weight predictor integrates learnable base 4D LUTs in a soft weighting strategy to enable adaptive Wavelet-prior 4D LUT generation for better video enhancement.

\begin{figure}[t]
        \centering
        \includegraphics[height=0.46\textwidth,width=0.47\textwidth]{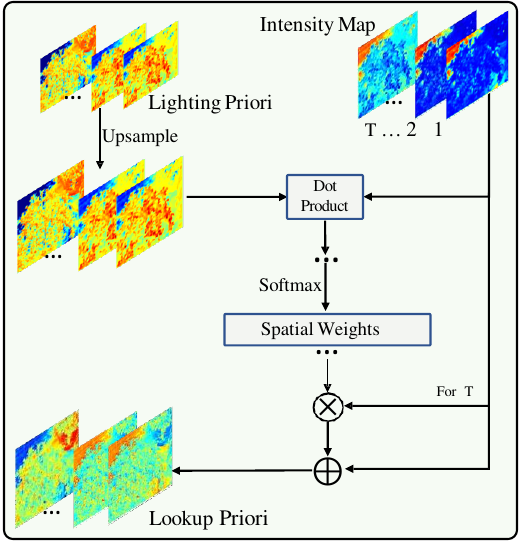}
        \caption{Workflow of the dynamic fusion strategy. The lighting prior generated in the wavelet low-frequency domain and the intensity map generated from the low-light video are weighted and summed for similarity to generate an optimized lookup priori.}
        \label{fig:4}
    \vspace{-1em}
    \end{figure}

\subsubsection{The Interpolation Step} Unlike 3D LUT, 4D LUT can store multiple colour spaces. Based on the original RGB values and the generated lookup prior, our proposed 4D LUT is indexed on a 4D space. Therefore for each input space, we use the mapping $\mu(\cdot)$ to obtain the stored value $L$ in the constructed Wavelet-prior 4D LUT, can be formulated as:
\begin{small}
\begin{equation}
L\{\hat{R}_{out},\hat{G}_{out},\hat{B}_{out}\}=\mu(R_{(i,j,k,e)} ,G_{(i,j,k,e)},B_{(i,j,k,e)}, E_{(i,j,k,e)}),
\tag{5}
\end{equation}
\end{small}
where $R_{(i,j,k,e)}, G_{(i,j,k,e)}, B_{(i,j,k,e)}, E_{(i,j,k,e)}$ denote the input red, green, blue, and enhancement a priori intensities, and $\hat{R}_{out},\hat{G}_{out},\hat{B}_{out}$ are the output color values of the mappings. We adopt $(x,y,z,s)=(\left \lfloor i \right\rfloor,\left \lfloor j \right \rfloor,\left \lfloor k \right \rfloor,\left \lfloor e \right \rfloor)$ to represent the sample point index of each mapping in the 4D LUT, where $\left \lfloor \cdot \right \rfloor$ represents the floor function. Furthermore, for the input $(R, G, B, E)$, which cannot be mapped to any sample point, we first locate the 16 nearest neighboring elements $\{(x,y,z,s),(x+1,y,z,s), \dots,(x+1,y+1,z+1,s+1)\}$ around the input index as sample points, and then subsequently apply quadrilinear interpolation to obtain the nearest sample points. Below we take the output red channel value $\hat{R}_{out}$ as an example:
\begin{equation}
\begin{split}
\hat{R}_{out} &= (1-O_r )\cdot(1-O_g )\cdot(1-O_b )\cdot(1-O_e )\cdot \hat{R}_{(x,y,z,s)} \\
&+O_r \cdot(1-O_g )\cdot(1-O_b )\cdot(1-O_e )\cdot \hat{R}_{(x+1,y,z,s)}\\
&+O_r\cdot(1-O_g )\cdot(1-O_b )\cdot(1-O_e )\cdot \hat{R}_{(x+1,y,z,s)}\\
&\quad \quad \quad \quad  \quad \quad \quad \quad \vdots \quad \\
&+O_r \cdot O_g \cdot O_b \cdot (1-O_e )\cdot \hat{R}_{(x+1,y+1,z+1,s)}\\
&+O_r \cdot O_g \cdot O_b \cdot O_e\cdot \hat{R}_{(x+1,y+1,z+1,s+1)},\\
\end{split}
\tag{6}
\end{equation}
where \(\{O_r, O_g, O_b, O_e\}\) denotes the offset from the input index \((i, j, k, e)\) to the defined sampling grid \((x, y, z, s)\), which satisfies \(\Delta \in [-1, 1]\) in each dimension. We show it as an example with \(O_r\):
\begin{equation}
O_r=[\frac{R_{(i,j,k,e)}-\hat{R}_{(x,y,z,s)}}{\hat{R}_{(x+1,y,z,s)}-\hat{R}_{(x,y,z,s)}},\frac{\hat{R}_{(x+1,y,z,s)}-R_{(i,j,k,e)}}{\hat{R}_{(x+1,y,z,s)}-R_{(x,y,z,s)}}].
\tag{7}
\end{equation}

\begin{figure}[t]
        \centering
        \includegraphics[height=0.4\textwidth,width=0.48\textwidth]{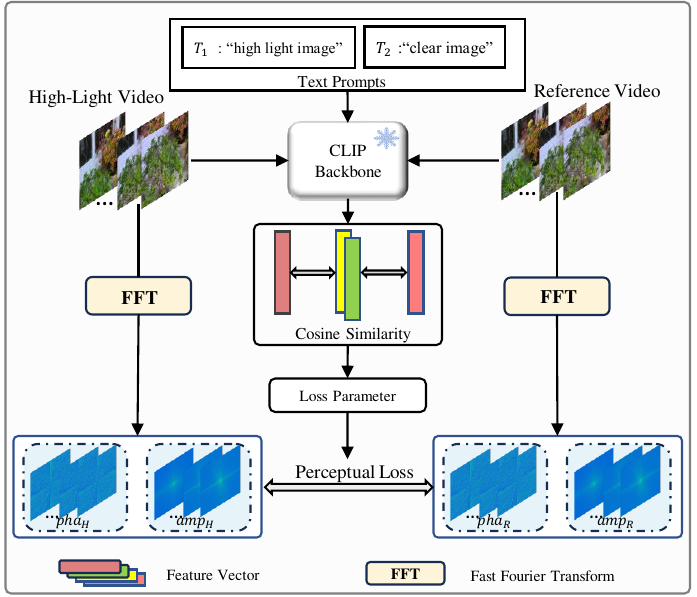}
        \caption{Detailed flow of text-driven appearance reconstruction. We first calculate the loss parameters using the frozen CLIP model. The phase and amplitude are obtained by performing the Fast Fourier Transform on the enhanced high-light video and the reference video, respectively, to construct the perceptual loss, and the loss parameters are introduced for dynamic adjustment.}
        \label{fig:D}
    \end{figure}

\subsubsection{Light-Enhanced 4D LUT Transformation} As shown in Fig. \ref{fig:7}, to better match the constructed Wavelet-prior 4D LUT, while enhancing the accuracy of the mapping. We first estimate the lighting prior in the wavelet low-frequency domain and a rough intensity map of the low-light video through a lightweight decoder. Subsequently, as shown in Fig. \ref{fig:4}, a dynamic fusion strategy is used to adaptively determine the spatial weights based on the correlation between the wavelet lighting prior and the target structure. Unlike static fusion strategies, such as simple summation, the dynamic fusion strategy can effectively combine the lighting prior with the video content to reduce the risk of alignment. Specifically, the strategy consists of three core steps: firstly, upsampling the lighting prior to match the size; secondly, obtaining the spatial weights based on the similarity computation of each frame of the intensity map of the video adaptively with the lighting prior; and thirdly, iteratively weighting and summing the similarities of the intensity maps of the low-light video. 

Finally, we perform lookup and interpolation in the Wavelet-prior 4D LUT based on the fused lookup prior and the RGB video to stabilise and visually friendly the enhanced low-light video. At the same time, we follow the setup of \cite{yang2022adaint,li2023fastllve} to perform the lookup operation using a binary search, which will facilitate the implementation of the lookup algorithm and reduce the time complexity due to the bounded and monotonically increasing nature of the generated sampling coordinates. The mapping results are also denoised \cite{chen2019real} to further refine the output. In particular, the enhanced dimension after the lighting priori processing greatly enhances the accuracy of the interpolation, which also provides an effective guarantee for maintaining the color consistency between frames.

\begin{table*}[t]\normalsize
\setlength{\tabcolsep}{20pt}
\renewcommand\arraystretch{1.3}
\caption{Quantitatively evaluated image-based and video-based baseline methods on the SDSD \cite{wang2021real} and SMID \cite{chen2019seeing} datasets, respectively. The best scores are shown by \textbf{highlighting}.}
\scalebox{0.9}{
\begin{tabular}{l|l|l|cc|cc}
\hline
\multirow{2}{*}{Type}  & \multirow{2}{*}{Models} & \multirow{2}{*}{Reference} & \multicolumn{2}{c|}{SDSD}      & \multicolumn{2}{c}{SMID}            \\ \cline{4-7} 
                       &                         &                            & PSNR↑          & SSIM↑          & PSNR↑          & SSIM↑               \\ \hline
\multirow{5}{*}{Image} & LLFlow \cite{wang2022low}                 & AAAI'22                    & 24.90          & 0.78          & 27.02          & 0.79                  \\
                       & SNRNet  \cite{xu2022snr}                & CVPR'22                    & 25.27          & 0.83          & 27.82          & 0.80                 \\
                       & PairLLE \cite{fu2023learning}                & CVPR'23                    & 21.76          & 0.68          & 22.73          & 0.63                  \\
                       & SMG-LLIE  \cite{xu2023low}              & CVPR'23                    & 26.49          & 0.84          & 27.41          & 0.78               \\
                       & CLIP-Lit \cite{liang2023iterative}               & ICCV'23                    & 21.96          & 0.73          & 18.64          & 0.62              \\
                       & NeRco  \cite{yang2023implicit}                 & ICCV'23                    & 20.07          & 0.66          & 19.91          & 0.64              \\
                        &FourierDiff  \cite{lv2024fourier}              & CVPR'24                    & 18.69          & 0.63          & 17.63         & 0.61               
                       \\ \hline
\multirow{9}{*}{Video} & MBLLVEN \cite{lv2018mbllen}               & BMVC'18                    & 21.79          & 0.66          & 22.67          & 0.68                 \\
                       & SMOID \cite{jiang2019learning}                  & ICCV'19                    & 23.45          & 0.69          & 23.64          & 0.71                 \\
                       & SMID \cite{chen2019seeing}                   & ICCV'19                    & 24.09          & 0.69          & 24.78          & 0.72              \\
                       & StableLLVE \cite{zhang2021learning}             & CVPR'21                    & 23.79          & 0.81          & 26.22          & 0.78                \\
                       & SDSDNet \cite{wang2021real}                 & ICCV'21                    & 24.92          & 0.73          & 26.03          & 0.75                  \\
                       & SGLLVE \cite{chhirolya2022low}               & BMVC'22                    & 23.46          & 0.79          & 24.72          & 0.70               \\
                       & LLVE-SEG \cite{liu2023low}               & AAAI'23                    & 25.89          & 0.76          & 25.66          & 0.75                  \\
                       & DP3DF \cite{xu2023deep}                  & AAAI'23                    & 26.03         & 0.79          & 27.39          & 0.77                 \\
                       & FASTLLVE \cite{li2023fastllve}               & ACM MM'23                  & 27.61          & 0.85          & 27.62          & 0.80                 \\ \hline
Video                  & Ours                    & \ \ \ \ \ \  -                        & \textbf{28.18} & \textbf{0.87} & \textbf{28.89} & \textbf{0.82} \\ \hline
\end{tabular}
}
\label{tab:1}
\end{table*}

\begin{figure}[t]
        \centering
        \includegraphics[height=0.4\textwidth,width=0.48\textwidth]{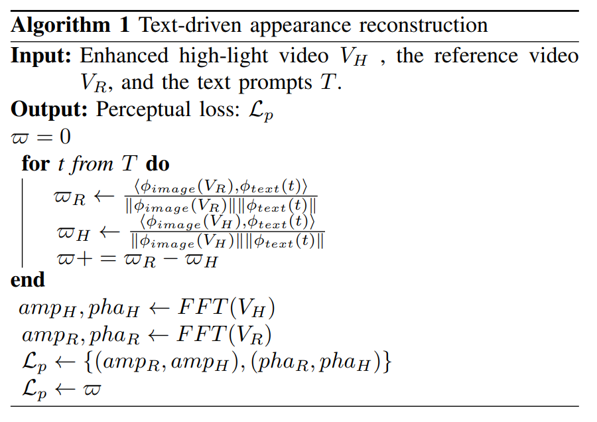}
        \label{fig:f}
\end{figure}

\subsection{Text-Driven Appearance Reconstruction}
Due to the great potential shown by the Contrast-Language-Image-Pre-Training (CLIP) model \cite{radford2021learning} in low-light image enhancement tasks \cite{liang2023iterative,xue2024low}, and also to further promote the enhancement results to keep the brightness consistent between frames and close to the reference video. Therefore, we dynamically adjust the enhancement process by introducing multimodal text and supervising the video reconstruction from both semantic and frequency domain levels in combination with Fourier spectra. To the best of our knowledge, this is the first time that multimodal text has been introduced to supervise the LLVE task. As shown in Fig. \ref{fig:D} and Algo. {\color{blue}1}, we first manually designed two text prompts, $T_1$ and $T_2$, i.e., high light image and clean image. Subsequently, the extracted high-light video frames are fed into the frozen CLIP model, which performs feature vector extraction via its two internal pre-trained text encoders $\phi_{text}$ and image encoders $\phi_{image}$. We compute the similarity between the image vectors and the text vectors to measure the difference between them, which can be expressed as follows:
\begin{equation}
D_S=\frac{\left \langle \phi_{image}(V),\phi_{text} (T ) \right \rangle}{\left \|  \phi_{image} (V) \right \|\left \| \phi_{text} (T) \right \| } ,
\tag{8}
\end{equation}
where $V$ denotes the input video frame and $T$ denotes the input text prompt. Therefore, the loss parameter $\varpi$ of the perceptual loss can be formulated as:
\begin{equation}
\begin{split}
\varpi =&\left \| D_S(V_R,T_1 )-D_S(V_H,T_1 ) \right \| \\
&+\left \| D_S(V_R,T_2 )-D_S(V_H,T_2 ) \right \| ,
\end{split}
\tag{9}
\end{equation}
where a larger $\varpi$ indicates a larger gap between the enhancement result and the reference video in the brightness prompts. Subsequently, we use the Fast Fourier Transform $(FFT)$ to construct the spatial spectrum, which consists of both amplitude ($amp$) and phase ($pha$) components. Among them, most of the brightness information is concentrated in the amplitude, while the structure and content information is closely related to the phase \cite{lv2024fourier}. Therefore, we dynamically adjust the weight between amplitude and phase according to the loss parameter $\varpi$ when constructing the perceptual loss $\mathcal{L}_{p}$ to optimise the brightness and content structure better:
\begin{equation}
amp_{H},pha_{H}=FFT(V_H),
\tag{10}
\end{equation}
\begin{equation}
amp_{R},pha_{R}=FFT(V_R),
\tag{11}
\end{equation}
\begin{equation}
\mathcal{L}_{p}=\varpi \parallel amp_H -amp_R \parallel_1 + (1-\varpi) \parallel pha_H - pha_R \parallel_1 .
\tag{12}
\end{equation}

\subsection{Model Training}
For the training optimization of the model, we first minimize the content difference between the output result $V_H$ and the reference video $V_R$ by combining the Charbonnier loss \cite{lai2017deep} and the SSIM loss \cite{wang2004image} as the content loss $\mathcal{L}_{c}$:
\begin{equation}
    \mathcal{L}_{c}=\sqrt{(V_H-V_R)^2+\epsilon^2} +\vartheta (1-SSIM(V_H,V_R)),
\tag{13}
\end{equation}
where $\epsilon$ is a very small constant to ensure the differentiability of the function. $\vartheta$ is a hyperparameter, which we empirically set to 0.1. 

 \begin{table}[t]\normalsize
\setlength{\tabcolsep}{15pt}
\renewcommand\arraystretch{1.3}
\caption{Quantitatively evaluated image-based and video-based baseline methods on the DID \cite{fu2023dancing} datasets. The best scores are shown by \textbf{highlighting}.}
\scalebox{0.9}{
\begin{tabular}{c|l|cc}
\hline
\multirow{2}{*}{Type}      & \multirow{2}{*}{Models} & \multicolumn{2}{c}{DID} \\ \cline{3-4} 
                           &                         & PSNR↑        & SSIM↑      \\ \hline
\multirow{3}{*}{Image}     
                           & SNRNet  \cite{xu2022snr}                & 24.05       & \textbf{0.90}      \\
                           & PairLLE  \cite{fu2023learning}               & 22.56       & 0.82      \\ 
                           & NeRco  \cite{yang2023implicit}               &  20.88     &  0.83    \\ \hline
\multirow{8}{*}{Video}     & MBLLVEN\cite{lv2018mbllen}                 & 24.82       & 0.89      \\
                           & SMOID \cite{jiang2019learning}                  & 22.57       & 0.87      \\
                           & SMID \cite{chen2019seeing}                   & 22.97       & 0.88      \\
                           & StableLLVE \cite{zhang2021learning}             & 21.64       & 0.83      \\
                           & SDSDNet \cite{wang2021real}                & 21.88       & 0.82      \\
                           & LLVE-SEG \cite{liu2023low}               & 23.85       & 0.86      \\
                           & DP3DF  \cite{xu2023deep}                 & 22.39       & 0.88      \\
                           & FASTLLVE \cite{li2023fastllve}               & 24.16       & 0.86      \\ \hline
\multicolumn{1}{l|}{Video} & Ours                    & \textbf{26.49}            & \ \textbf{0.90 }     \\ \hline
\end{tabular}
}
\label{tab:2}
\end{table}

Since previous LUT-based methods have demonstrated the effectiveness of regularisation loss \cite{zeng2020learning,yang2022adaint,li2023fastllve} by smoothing loss $\mathcal{L}_{s}$ and monotonicity loss $\mathcal{L}_{m}$. Therefore, in this paper, we also constrain the output of the 4D LUT and the generation of artefacts by two regularisation losses to ensure the stability and robustness of the mapping space:
\begin{equation}
\begin{split}
\mathcal{L}_{s}= \sum_{C\in\{\hat{R},\hat{G},\hat{B}\}}^{i\in \{x,y,z,s\}} \sum_{0}^{N-1}\left \| C_{i+1}-C_{i} \right \|^2
+\sum_{1}^{3} \left \| w_n \right \| ^2,
\end{split}
\tag{14}
\end{equation}

\begin{equation}
\begin{split}
\mathcal{L}_{m}= \sum_{C\in\{\hat{R},\hat{G},\hat{B}\}}^{i\in \{x,y,z,s\}} \sum_{0}^{N-1}Relu( C_{i}-C_{i+1} ),
\end{split}
\tag{15}
\end{equation}
where $C_{i+1}$ and $C_{i}$ are the mapped output red, green, and blue colors corresponding to the defined sampling point in 4D LUT. $w_n$ denotes the video-related weights output from the weight predictor. Relu denotes the ReLU activation function.

As mentioned above, our total loss during training can be summarised as:
\begin{equation}
\mathcal{L}_{total}= \mathcal{L}_{c}+\mathcal{L}_{p}+\beta _s\mathcal{L}_{s}+\beta _m\mathcal{L}_{m},
\tag{16}
\end{equation}
where $\beta _s$ and $\beta _m$ are hyperparameters for balancing the size of network losses, we set to $1\times 10^{-4}$ and 10, respectively.

\section{EXPERIMENTS} \label{EXPERIMENTS}
\subsection{Experimental Settings}
\begin{figure*}[t]
        \centering
        \includegraphics[height=0.4\textwidth,width=\textwidth]{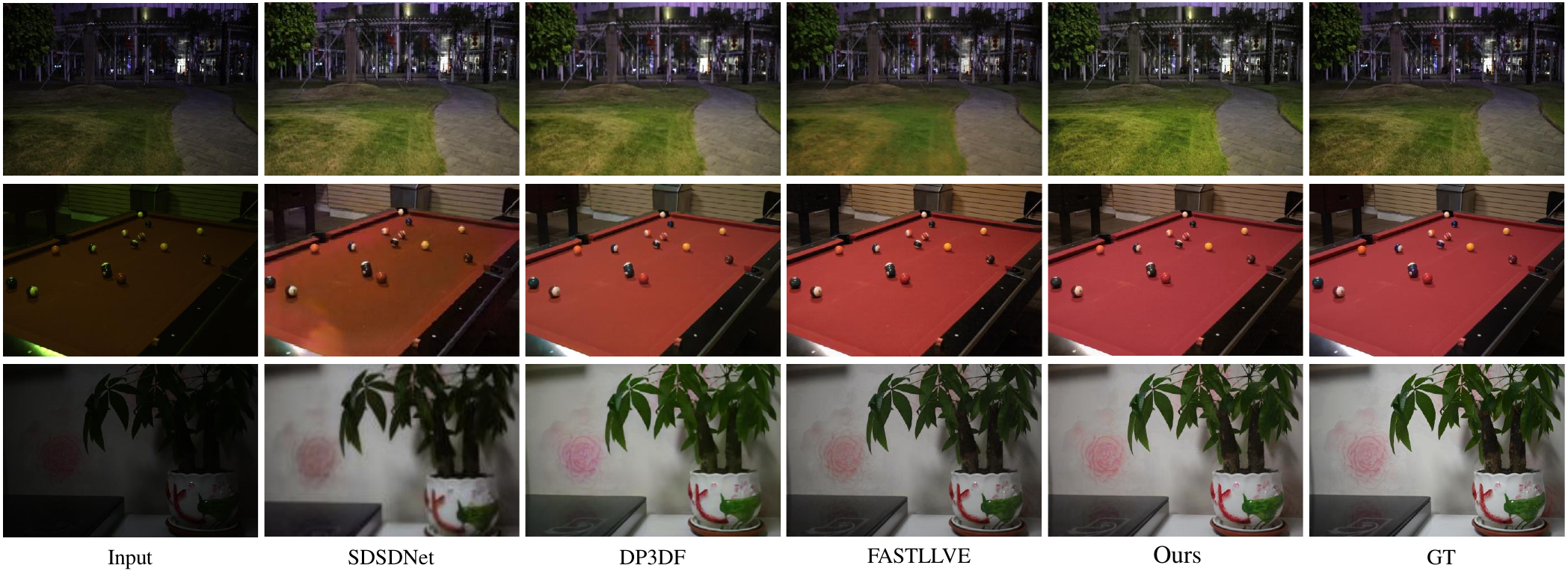}
        \caption{Visual comparison of our method with State-of-the-art methods on the SDSD \cite{wang2021real} (row 1), SMID \cite{chen2019seeing} (row 2), and DID \cite{fu2023dancing} (row 3) datasets. Our method is closer to a reference video, best viewed by zooming in. }
        \label{fig:5}
    \end{figure*}

\begin{figure*}[t]
        \centering
        \includegraphics[height=0.28\textwidth,width=\textwidth]{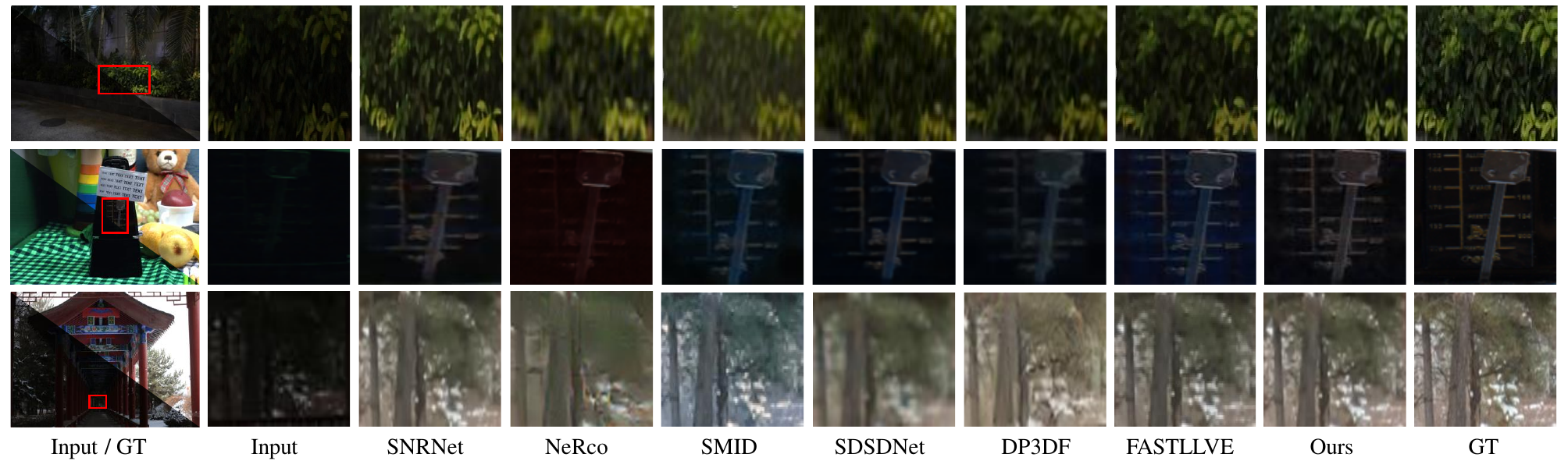}
        \caption{Comparison of video details between our method and the baseline method performed on the  SDSD \cite{wang2021real}(row 1), SMID \cite{chen2019seeing}(row 2), and DID \cite{fu2023dancing}(row 3) datasets. Best viewed by zooming in.}
        \label{fig:6}
    \end{figure*}

\subsubsection{Dataset} Our network is first evaluated on three publicly available benchmark datasets, SMID \cite{chen2019seeing}, SDSD \cite{wang2021real}, and DID \cite{fu2023dancing}, which contain various real-world videos with different motion patterns and degradations. Specifically, SMID consists of still videos containing reference videos obtained through long-time exposures with $18278$ and $1470$ training and test video frames, respectively. SDSD is a dynamic video dataset collected through electromechanical devices, including indoor and outdoor subsets, and contains $23542$ and $750$ training and test video frames. DID is a dataset of real-world videos with significant spatial movement, strict spatial alignment and diverse scene content of the dynamic video dataset with $413$ video pairs and a total of $41038$ frames. Finally, we also further test the model with low-light videos from the DSEC dataset \cite{gehrig2021dsec} to verify its generalisation ability.  

For the evaluation metrics, we propose to use two full-reference distortion metrics, PNSR and SSIM \cite{wang2004image}, to evaluate the performance of our method. In addition, for the DSEC dataset, we used three non-reference perceptual metrics: NIQE \cite{mittal2012making}, BRISQUE \cite{mittal2012no}, and PI \cite{blau20182018} to evaluate the visual quality of the enhancement results. The lower the metrics, the better the visual quality.

\subsubsection{Implementation Details} We implemented our framework using PyTorch on a single NVIDIA RTX 3090 GPU. The network uses the Adam optimiser with the initial learning rate set to $4\times10^{-4}$ and the batch size and training size set to $8$ and $256\times256$. Also, the lightweight encoder/decoder consists of five 3D convolutional blocks/deconvolutional blocks with a convolutional kernel set to $3\times3\times3$.

\subsubsection{Comparison Methods} To verify the effectiveness of the method proposed in this paper, we compared it with the State-of-the-art methods in recent years. This includes the video-based methods MBLLEN \cite{lv2018mbllen}, SMID \cite{chen2019seeing}, SMOID \cite{jiang2019learning}, SDSDNet \cite{wang2021real}, StableLLVE \cite{zhang2021learning}, SGLLVE \cite{chhirolya2022low}, LLVE-SEG \cite{liu2023low}, DP3DF \cite{xu2023deep}, and FASTLLVE \cite{li2023fastllve}. Image-based methods: LLFlow \cite{wang2022low}, SNRNet \cite{xu2022snr}, PairLLE \cite{fu2023learning}, SMG-LLIE \cite{xu2023low}, CLIP-Lit \cite{liang2023iterative}, NeRco \cite{yang2023implicit}, and FourierDiff \cite{lv2024fourier}.

\begin{table}[t]\normalsize
\setlength{\tabcolsep}{8pt}
\renewcommand\arraystretch{1.2}
\caption{Quantitatively evaluated image-based and video-based baseline methods on the DSEC \cite{gehrig2021dsec} datasets. The best scores are shown by \textbf{highlighting}.}
\scalebox{0.9}{
\begin{tabular}{c|l|ccc}
\hline
Type                   & Models     & NIQE↓ & BRISQUE↓ & PI↓   \\ \hline
\multirow{3}{*}{Image} & SNRNet\cite{xu2022snr}     & 3.82 & 30.96   & 3.86 \\
                       & CLIP-Lit\cite{liang2023iterative}   & 5.09 & 24.93   & 3.47 \\
                       & NeRco\cite{yang2023implicit}      & 3.68 & 25.40   & 2.92 \\ \hline
\multirow{7}{*}{Video} & MBLLVEN\cite{lv2018mbllen}    & 4.77 & 32.17   & 3.99 \\
                       & SMID\cite{chen2019seeing}        & 5.73 & 37.43   & 4.75 \\
                       & StableLLVE \cite{zhang2021learning} & 4.79 & 32.12   & 4.87 \\
                       & SDSDNet \cite{wang2021real}     & 5.39 & 25.72   & 5.28 \\
                       & SGLLVE \cite{chhirolya2022low}     & 4.35 & 25.93   & \textbf{2.81} \\
                       & DP3DF \cite{xu2023deep}     & 3.89 & 30.81   & 4.52 \\
                       & FASTLLVE \cite{li2023fastllve}  & 3.90 & 26.10   & 3.15 \\ \hline
Video                  & Ours       & \textbf{3.66} & \textbf{24.84}   & 3.09 \\ \hline
\end{tabular}
}
\label{tab:3}
\end{table}


\begin{figure*}[t]
        \centering
        \includegraphics[height=0.14\textwidth,width=\textwidth]{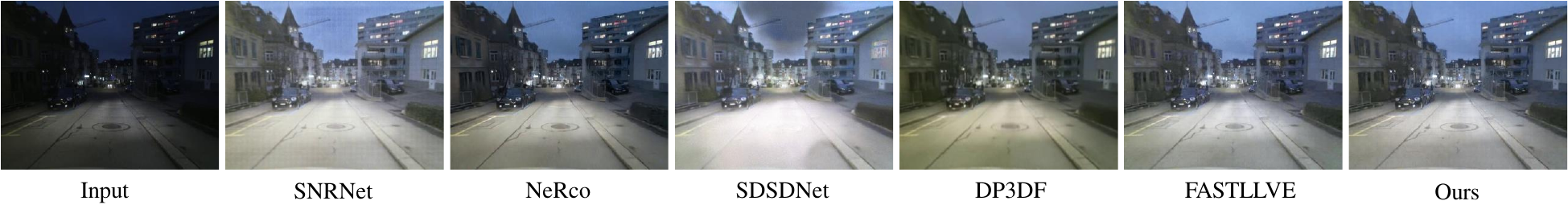}
        \caption{Visual comparison of our method with the baseline method performed on the DSEC \cite{gehrig2021dsec} dataset. }
        \label{fig:8}
    \end{figure*}

\subsection{Experimental Results}
\subsubsection{Quantitative Comparison} As shown in Table \ref{tab:1}, we show the evaluation results of all competing methods on the SDSD \cite{wang2021real} and SMID \cite{chen2019seeing} datasets. It is easy to see that our method obtains the highest PSNR and SSIM evaluation scores on both datasets, especially in PSNR evaluation. Specifically, in the SDSD dataset, we obtained a PSNR improvement of 0.57 dB (=28.18-27.61) and an SSIM increase of 0.02 (=0.87-0.85) compared to the second-ranked FASTLLVE. In the SMID dataset, we also acquired a 1.07 dB (=28.89-27.82) enhancement and 0.02 (=0.82-0.80) SSIM rise compared to the second SNRNet. Meanwhile, Table \ref{tab:2} demonstrates the quantitative results in the DID \cite{fu2023dancing} dataset. On a dynamic video dataset like DID, which has a large spatial offset, we have a clear advantage with a PSNR boost of 1.67 dB (=26.49-24.82). This further confirms the potential of our method in enhancing real-world videos. In addition, our cross-dataset validation on the DSEC \cite{gehrig2021dsec} dataset, as shown in Table \ref{tab:3}, also fully confirms the generalization ability of the model. We obtained the best scores for both NIQE and BRISQUE among the three non-reference metrics evaluated.

\subsubsection{Visual Comparison} Fig. \ref{fig:5} illustrates the visual effect between our and baseline methods. Where the images in rows 1-3 are selected from the SDSD, SMID, and DID test sets, respectively. It can be intuitively seen that the results enhanced by our method have more friendly visual effects, including accurate colors, balanced brightness, and enhanced contrast. In comparison, the video-based method SDSDNet has more blurred artifacts and details. DP3DF has more noise and less desirable contrast. This indicates that current methods cannot better generalize the correlation between video frames when dealing with video tasks. Compared with FastLLVE, which is also a lookup table-based method, our colors have better contrast, demonstrating that our constructed wavelet prior effectively compensates for the loss of the lookup prior in undesirable environments.

In addition, to further validate the performance of the proposed method and the processing of details, as shown in Fig. \ref{fig:6}, we demonstrate the effect of WaveLUT with various baseline methods in detail processing. It is easy to see that the image-based methods will have more color bias. The video-based methods SMID and SDSDNet have unnatural colors and lead to visual blurring. DP3DF and FASTLLVE have more artifacts, thus leading to unsatisfactory visual perception. In contrast, our proposed method is closer to the color of the reference video, which further proves the accuracy of WaveLUT color mapping. Finally, Fig. \ref{fig:8} shows the qualitative results of the cross-dataset tests performed on the DSEC dataset.

\begin{table}[]
\caption{Ablation studies of the designed model components.}
\renewcommand\arraystretch{1.3}
\scalebox{0.8}{
\begin{tabular}{lccc|cc}
\hline
Methods                        & Wavelet-prior 4D LUT & Lighting Priori & Dynamic Fusion & PSNR  & SSIM \\ \hline
\multicolumn{1}{l|}{\#1}     & \usym{2613}     & \usym{2613}             & \usym{2613}             
& 24.46 & 0.86 \\
\multicolumn{1}{l|}{\#2}     & \usym{2613}                 & \usym{1F5F8}            & \usym{1F5F8}              
& 25.69 & 0.88 \\
\multicolumn{1}{l|}{\#3}     & \usym{1F5F8}                     & \usym{2613}               & \usym{2613}              
& 26.07. & 0.89 \\
\multicolumn{1}{l|}{\#4}     & \usym{1F5F8}                    & \usym{1F5F8}                & \usym{2613}              & 26.21 & 0.89 \\
\multicolumn{1}{l|}{default} & \usym{1F5F8}                     & \usym{1F5F8}                &\usym{1F5F8}               & \textbf{26.49} &  \textbf{0.90}\\ \hline
\end{tabular}
}
\label{t1}
\end{table}

\subsection{Ablation Study}
To assess the effectiveness of the proposed components, we performed ablation studies on different components.
\subsubsection{Model design}
The positive effect of our model can be easily verified from Table \ref{t1}.  Where '\#1' indicates the base setup without any additions. '\#2' indicates that no Wavelet-prior 4D LUT is constructed, and only the lighting prior and dynamic fusion strategy are used. '\#3' denotes constructing Wavelet-prior 4D LUT only. and '\#4' denotes not using dynamic fusion strategy. 'default' represents the complete network setup. It can be seen that ‘\#2’ has a PSNR improvement of 1.23 dB by compensating for the lookup prior, which verifies that the constructed prior can be corrupted to some extent in adverse environments, resulting in biased mappings. In contrast, ‘\#3’ has a PSNR improvement of 1.61 dB on the network performance by constructing a Wavelet-prior 4D LUT, which validates the effectiveness of constructing the 4D LUT via the wavelet transform. Meanwhile, the comparison of '\#3' and '\#4' with 0.14 dB PSNR enhancement also highlights that the combination of Wavelet-prior 4D LUT and wavelet transform-based lookup priori can optimize the underlying color of the LUT and the accuracy of the mapping even further. Through the complete network setup, we verify that the Wavelet-prior 4D LUT can achieve a large performance improvement with the effective cooperation of the lighting priori and the dynamic fusion strategy, further proving the importance of the dynamic fusion strategy.

\begin{table}[]
\caption{Ablation studies of the loss function terms.}
\centering
\renewcommand\arraystretch{1.2}
\scalebox{1.1}{
\begin{tabular}{l|ccccc}
\hline
\rowcolor[HTML]{EFEFEF} 
Methods & w/o $\mathcal{L}_{c}$ & w/o $\mathcal{L}_{p}$ & w/o $\mathcal{L}_{s} $ & w/o $\mathcal{L}_{m}$ & default \\ \hline
PSNR   & 23.88  & 28.53  &  28.72      &28.63        & \textbf{28.89}   \\
SSIM   & 0.73   & 0.82   &  0.82      & 0.82       &\textbf{ 0.82}    \\ \hline
\end{tabular}
}
\label{loss}
\end{table}


\subsubsection{Loss function}
To verify the effectiveness of each loss function, we performed ablation experiments on them on the SMID dataset. The experimental results are shown in Table \ref{loss}. $\mathcal{L}_{c}$ denotes the reconstructed content loss. $\mathcal{L}_{p}$ denotes the semantically guided perceptual loss. $\mathcal{L}_{s}$ and $\mathcal{L}_{m}$ denote the smooth regularization loss and monotone regularization loss, respectively. It can be intuitively seen that the lack of $\mathcal{L}_{c}$ decreases PSNR and SSIM substantially (5.01 dB and 0.09), which illustrates the importance of $\mathcal{L}_{c}$ to the video reconstruction process, which effectively constrains the enhancement results from generating unwanted content and approaching the reference video. With the introduction of $\mathcal{L}_{p}$, the PSNR is increased by 0.36 dB, which verifies that the perceptual loss can maintain a dynamic balance between luminance and content during the enhancement process, and confirms the effectiveness of the designed text-driven appearance reconstruction method. By increasing $\mathcal{L}_{s}$, the PSNR can be improved from 28.72 dB to 28.89 dB, which proves that the smooth regularisation loss can guarantee the color mapping from the input four-dimensional space (RGBE) to the target three-dimensional space (RGB). When $\mathcal{L}_{m}$ is introduced, the color transform maintains the relative brightness/saturation of the colors while covering the entire RGBE space, and the PSNR performance is improved by 0.26 dB.

\section{Conclusions} \label{Conclusions}
In this paper, we propose WaveLUT for efficient implementation of low-light video enhancement. It effectively improves the mapping accuracy and color consistency between video frames while maintaining high efficiency. In addition, to further optimize the lookup prior, we design a dynamic fusion strategy to fuse different prior knowledge, which further ensures the brightness and contrast between video frames. In the training phase, we also propose a text-driven appearance reconstruction method, which effectively combines Fourier spectrum and multimodal semantics to optimize the enhancement results further.

Despite the excellent performance and visual perception of our proposed WaveLUT, the method still has limitations and objectives that need further exploration. Firstly, our method improves the mapping accuracy by optimizing the lookup prior, but for very dark scenes with limited cues, the construction of the lookup prior is challenging, and thus the possibility of color mapping bias also exists. In future work, we intend to explore further the properties of very dark scenes based on WaveLUT to construct more accurate and faster video enhancement algorithms.

\bibliographystyle{IEEEtran}
\bibliography{mybibfile}

\vfill

\end{document}